\pdfoutput=1
\documentclass[11pt,a4paper]{article}
\usepackage[table,xcdraw]{xcolor}
\usepackage[]{acl2020}
\usepackage{times}
\usepackage{latexsym}

\usepackage{graphicx}
\usepackage{algorithm}
\usepackage{algorithmic}
\usepackage{multirow}
\usepackage{tabularx}
\usepackage{booktabs}
\usepackage{amssymb}

\usepackage{microtype}

\aclfinalcopy 
\def\aclpaperid{1804} 

\title{GoEmotions: A Dataset of Fine-Grained Emotions}

\author{{\bf Dorottya Demszky}\textsuperscript{1}\thanks{\quad Work done while at Google Research.}\quad {\bf Dana Movshovitz-Attias}\textsuperscript{2}\quad {\bf Jeongwoo Ko}\textsuperscript{2}\vspace{1mm}\\{\bf Alan Cowen}\textsuperscript{2}\quad {\bf Gaurav Nemade}\textsuperscript{2}\quad {\bf Sujith Ravi}\textsuperscript{3}\textsuperscript{*}\vspace{1mm}\\\textsuperscript{1}Stanford Linguistics\quad \textsuperscript{2}Google Research\quad\textsuperscript{3}Amazon Alexa\vspace{1mm}\\ \texttt{ddemszky@stanford.edu}\\\texttt{\{danama, jko, acowen, gnemade\}@google.com}\\\texttt{sravi@sravi.org}}

\date{}

\date{}

\begin{document}
\maketitle
\begin{abstract}
  

Understanding emotion expressed in language has a wide range of applications, from building empathetic chatbots to detecting harmful online behavior.  Advancement in this area can be improved using large-scale datasets with a fine-grained typology, adaptable to multiple downstream tasks. We introduce GoEmotions, the largest manually annotated dataset of 58k English Reddit comments, labeled for 27 emotion categories or Neutral. We demonstrate the high quality of the annotations via Principal Preserved Component Analysis. We conduct transfer learning experiments with existing emotion benchmarks to show that our dataset generalizes well to other domains and different emotion taxonomies. Our BERT-based model achieves an average F1-score of .46 across our proposed taxonomy, leaving much room for improvement.\footnote{Data and code available at \url{https://github.com/google-research/google-research/tree/master/goemotions}.}
\end{abstract}

\section{Introduction}

Emotion expression and detection are central to the human experience and social interaction.
With as many as a handful of words we are able to express a wide variety of subtle and complex emotions, and it has thus been a long-term goal to enable machines to understand affect and emotion \citep{picard1997affective}.

In the past decade, NLP researchers made available several datasets for language-based emotion classification for a variety of domains and applications, including for news headlines \citep{affective-text-2007-semeval}, tweets \citep{crowdflower-2016,mohammad-etal-2018-semeval}, and narrative sequences \citep{liu2019dens}, to name just a few. However, existing available datasets are (1) mostly small, containing up to several thousand instances, and (2) cover a limited emotion taxonomy, with coarse classification into Ekman \citep{ekman1992argument} or Plutchik \citep{plutchik1980general} emotions.

Recently, \citet{unified-klinger2018analysis} have aggregated 14 popular emotion classification corpora under a unified framework that allows direct comparison of the existing resources. 
Importantly, their analysis suggests annotation quality gaps in the largest manually annotated emotion classification dataset, \citet{crowdflower-2016}, containing 40K tweets labeled for one of 13 emotions. While their work enables such comparative evaluations, it highlights the need for a large-scale, consistently labeled emotion dataset over a fine-grained taxonomy, with demonstrated high-quality annotations.

\begin{table}[t]
\centering
    
    \resizebox{\linewidth}{!}{

    \begin{centering}
\begin{tabular}{@{}ll@{}}
\toprule
\textbf{Sample Text}                                                                                              & \textbf{Label(s)}                                             \\ \midrule
\begin{tabular}[c]{@{}l@{}}OMG, yep!!! That is the final ans-\\wer. Thank you so much!\end{tabular}         & \begin{tabular}[c]{@{}l@{}}gratitude,\\ approval\end{tabular} \\ \midrule
\begin{tabular}[c]{@{}l@{}}I'm not even sure what it is, why\\do  people hate it\end{tabular}              & confusion                                                     \\ \midrule
Guilty of doing this tbph              & remorse                                                     \\ \midrule
\begin{tabular}[c]{@{}l@{}}This caught me off guard for real.\\ I'm actually off my bed laughing \end{tabular}              & \begin{tabular}[c]{@{}l@{}}surprise,\\ amusement\end{tabular}                                                   \\ \midrule
\begin{tabular}[c]{@{}l@{}}I tried to send this to a friend but\\ {[}NAME{]} knocked it away.\end{tabular} & disappointment                                                \\ \bottomrule
\end{tabular}
    \end{centering}
 }
    \caption{Example annotations from our dataset.}
    \label{tab:intro_examples}
\end{table}

To this end, we compiled GoEmotions, the largest human annotated dataset of 58k carefully selected Reddit comments, labeled for 27 emotion categories or Neutral, with comments extracted from popular English subreddits. Table~\ref{tab:intro_examples} shows an illustrative sample of our collected data. We design our emotion taxonomy considering related work in psychology and coverage in our data. In contrast to Ekman's taxonomy, which includes only one positive emotion (joy), our taxonomy includes a large number of positive, negative, and ambiguous emotion categories, making it suitable for downstream conversation understanding tasks that require a subtle understanding of emotion expression, such as the analysis of customer feedback or the enhancement of chatbots.

We include a thorough analysis of the annotated data and the quality of the annotations. Via Principal Preserved Component Analysis \citep{cowen2019primacy}, we show a strong support for reliable dissociation among all 27 emotion categories, indicating the suitability of our annotations for building an emotion classification model.

We perform hierarchical clustering  on the emotion judgments, finding that emotions related in intensity cluster together closely and that the top-level clusters correspond to sentiment categories. These relations among emotions allow for their potential grouping into higher-level categories, if desired for a downstream task.


We provide a strong baseline for modeling fine-grained emotion classification over GoEmotions. By fine-tuning a BERT-base model \citep{devlin2019bert}, we achieve an average F1-score of .46 over our taxonomy, .64 over an Ekman-style grouping into six coarse categories and .69 over a sentiment grouping. These results leave much room for improvement, showcasing this task is not yet fully addressed by current state-of-the-art NLU models.


We conduct transfer learning experiments with existing emotion benchmarks to show that our data can generalize to different taxonomies and domains, such as tweets and personal narratives. Our experiments demonstrate that given limited resources to label additional emotion classification data for specialized domains, our data can provide baseline emotion understanding and contribute to increasing model accuracy for the target domain.

\section{Related Work}
\label{sec:related_work}

\subsection{Emotion Datasets}
Ever since Affective Text  \citep{affective-text-2007-semeval}, the first benchmark for emotion recognition was introduced, the field has seen several emotion datasets that vary in size, domain and taxonomy  \citep[cf.][]{unified-klinger2018analysis}. The majority of emotion datasets are constructed manually, but tend to be relatively small. 
The largest manually labeled dataset is \citet{crowdflower-2016}, with 39k labeled examples, which were found by \citet{unified-klinger2018analysis} to be noisy in comparison with other emotion datasets. 
Other datasets are automatically weakly-labeled, based on emotion-related hashtags on Twitter \citep{wang2012harnessing, abdul2017emonet}. We build our dataset manually, making it the largest human annotated dataset, with multiple annotations per example for quality assurance.

Several existing datasets come from the domain of Twitter, given its informal language and expressive content, such as emojis and hashtags. Other datasets annotate news headlines \citep{affective-text-2007-semeval}, dialogs \citep{li2017dailydialog}, fairy-tales \citep{alm-etal-2005-emotions}, movie subtitles \citep{ohman2018creating}, sentences based on FrameNet \citep{ghazi2015detecting}, or self-reported experiences \citep{scherer1994evidence} among other domains. We are the first to build on Reddit comments for emotion prediction.


\subsection{Emotion Taxonomy}
\label{ssec:related_work_emotion_taxonomy}
 One of the main aspects distinguishing our dataset is its emotion taxonomy. The vast majority of existing datasets contain annotations for minor variations of the 6 basic emotion categories (joy, anger, fear, sadness, disgust, and surprise) proposed by \citet{ekman1992there} and/or along affective dimensions (valence and arousal) that underpin the circumplex model of affect \cite{russell2003core,buechel2017emobank}. 
 

Recent advances in psychology have offered new conceptual and methodological approaches to capturing the more complex ``semantic space'' of emotion \citep{cowen2019mapping} by studying the distribution of emotion responses to a diverse array of stimuli via computational techniques. Studies guided by these principles have identified 27 distinct varieties of emotional experience conveyed by short videos \cite{cowen2017self}, 13 by music \cite{cowen2019music}, 28 by facial expression \cite{cowen2019face}, 12 by speech prosody \cite{cowen2019primacy}, and 24 by nonverbal vocalization \cite{cowen2018mapping}. In this work, we build on these methods and findings to devise our granular taxonomy for text-based emotion recognition and study the dimensionality of language-based emotion space.

\subsection{Emotion Classification Models}
Both feature-based and neural models have been used to build automatic emotion classification models. Feature-based models often make use of hand-built lexicons, such as the Valence Arousal Dominance Lexicon \citep{mohammad2018obtaining}. Using representations from BERT \citep{devlin2019bert}, a transformer-based model with language model pre-training, has recently shown to reach state-of-the-art performance on several NLP tasks, also including emotion prediction: the top-performing models in the EmotionX Challenge \citep{hsu-ku-2018-socialnlp} all employed a pre-trained BERT model. We also use the BERT model in our experiments and we find that it outperforms our biLSTM model.
\section{GoEmotions}
\label{sec:data_collection}

Our dataset is composed of 58K Reddit comments, labeled for one or more of 27 emotion(s) or Neutral.

\subsection{Selecting \& Curating Reddit comments}

We use a Reddit data dump originating in the reddit-data-tools project\footnote{\url{https://github.com/dewarim/reddit-data-tools}}, which contains comments from 2005 (the start of Reddit) to January 2019. We select subreddits with at least 10k comments and remove deleted and non-English comments.

Reddit is known for a demographic bias leaning towards young male users \citep{duggan20136}, which is not reflective of a globally diverse population. 
The platform also introduces a skew towards toxic, offensive language \citep{mohan2017impact}. Thus, Reddit content has been used to study depression \citep{pirina2018identifying}, microaggressions \citep{breitfeller2019finding}, 
and \citet{Yanardag2019norman} have shown the effect of using biased Reddit data by training a ``psychopath" bot. To address these concerns, and enable building broadly representative emotion models using GoEmotions, we take a series of data curation measures to ensure our data does not reinforce general, nor emotion-specific, language biases.

We identify harmful comments using pre-defined lists  containing offensive/adult, vulgar (mildly offensive profanity), identity, and religion terms (included as supplementary material). These are used for data filtering and masking, as described below.
Lists were internally compiled and we believe they are comprehensive and widely useful for dataset curation, however, they may not be complete.

\paragraph{Reducing profanity.}  We remove subreddits that are not safe for work\footnote{\url{http://redditlist.com/nsfw}} and where 10\%+ of comments include offensive/adult and vulgar tokens. We remove remaining comments that include offensive/adult tokens. Vulgar comments are preserved as  we believe they are central to learning about negative emotions. The dataset includes the list of filtered tokens.

\paragraph{Manual review.} We manually review identity comments and remove those offensive towards a particular ethnicity, gender, sexual orientation, or disability, to the best of our judgment.

\paragraph{Length filtering.} We apply NLTK's word tokenizer and select comments 3-30 tokens long, including punctuation. To create a relatively balanced distribution of comment length, we perform downsampling, capping by the number of comments with the median token count (12).

\paragraph{Sentiment balancing.} We reduce sentiment bias by removing subreddits with little representation of positive, negative, ambiguous, or neutral sentiment. To estimate a comment's sentiment, we run our emotion prediction model, trained on a pilot batch of 2.2k annotated examples. The mapping of emotions into sentiment categories is found in Figure~\ref{fig:hierarchical_corr}. We exclude subreddits consisting of more than 30\% neutral comments or less than 20\% of negative, positive, or ambiguous comments.

\paragraph{Emotion balancing.}
We assign a predicted emotion to each comment using the pilot model described above. Then, we reduce emotion bias by downsampling the weakly-labelled data, capping by the number of comments belonging to the median emotion count.

\paragraph{Subreddit balancing.} To avoid over representation of popular subreddits, we perform downsampling, capping by the median subreddit count. 

From the remaining 315k comments (from 482 subreddits), we randomly sample for annotation.

\paragraph{Masking.} We mask proper names referring to people with a [NAME] token, using a BERT-based Named Entity Tagger~\citep{tsai2019small}. We mask religion terms with a [RELIGION] token. The list of these terms is included with our dataset. Note that raters viewed unmasked comments during rating.

\subsection{Taxonomy of Emotions}
\label{ssec:taxonomy}
When creating the taxonomy, we seek to jointly maximize the following objectives.

1. \emph{Provide greatest coverage in terms of emotions expressed in our data.} 
To address this, we manually labeled a small subset of the data, and ran a pilot task where raters can suggest emotion labels on top of the pre-defined set. 

2. \emph{Provide greatest coverage in terms of kinds of emotional expression.}
We consult psychology literature on emotion expression and recognition \citep{plutchik1980general, cowen2017self,cowen2019primacy}. Since, to our knowledge, there has not been research that identifies principal categories for emotion recognition in the domain of text (see Section~\ref{ssec:related_work_emotion_taxonomy}), we consider those emotions that are identified as basic in other domains (video and speech) and that we can assume to apply to text as well.

3. \emph{Limit overlap among emotions and limit the number of emotions.}
We do not want to include emotions that are too similar, since that makes the annotation task more difficult. Moreover, combining similar labels with high coverage would result in an explosion in annotated labels.

The final set of selected emotions is listed in Table~\ref{tab:results}, and Figure~\ref{fig:emo_distr}. See Appendix~\ref{sec:appendix_taxonomy_selection} for more details on our multi-step taxonomy selection procedure.



\subsection{Annotation}
\label{ssec:annotation}

We assigned three raters to each example. For those examples where no raters agree on at least one emotion label, we assigned two additional raters. All raters are native English speakers from India.\footnote{\citet{cowen2019primacy} find that emotion judgments in Indian and US English speakers largely occupy the same dimensions.}

\paragraph{Instructions.} Raters were asked to identify the emotions expressed by the writer of the text, given pre-defined emotion definitions (see Appendix~\ref{sec:emotion_definitions}) and a few example texts for each emotion. Raters were free to select multiple emotions, but were asked to only select those ones for which they were reasonably confident that it is expressed in the text. If raters were not certain about any emotion being expressed, they were asked to select Neutral. We included a checkbox for raters to indicate if an example was particularly difficult to label, in which case they could select no emotions. We removed all examples for which no emotion was selected.

\paragraph{The rater interface.} 
Reddit comments were presented with no additional metadata (such as the author or subreddit). To help raters navigate the large space of emotion in our taxonomy, they were presented a table containing all emotion categories aggregated by sentiment (by the mapping in Figure~\ref{fig:hierarchical_corr}) and whether that emotion is generally expressed towards something (e.g. disapproval) or is more of an intrinsic feeling (e.g. joy). The instructions highlighted that this separation of categories was by no means clear-cut, but captured general tendencies, and we encouraged raters to ignore the categorization whenever they saw fit. Emotions with a straightforward mapping onto emojis were shown with an emoji in the UI, to further ease their interpretation.


\section{Data Analysis} \label{sec:data_analysis}

\begin{table}[t!]
    
    \resizebox{\linewidth}{!}{

    \begin{centering}
\begin{tabular}{|
>{\columncolor[HTML]{EFEFEF}}l l|}
\hline
\textbf{Number of examples}                                                                                               & 58,009       \\ \hline
\textbf{Number of emotions}                                                                                               & 27 + neutral \\ \hline
\textbf{Number of unique raters}                                                                                          & 82           \\ \hline
\textbf{Number of raters / example}                                                                                       & 3 or 5         \\ \hline
\textbf{\begin{tabular}[c]{@{}l@{}}Marked unclear or\\ difficult to label\end{tabular}}                                   & 1.6\%   \\ \hline
\cellcolor[HTML]{EFEFEF}                                                                                                  & 1: 83\%     \\
\cellcolor[HTML]{EFEFEF}                                                                                                  & 2: 15\%         \\
\cellcolor[HTML]{EFEFEF}                                                                                                  & 3: 2\%        \\
\multirow{-4}{*}{\cellcolor[HTML]{EFEFEF}\textbf{Number of labels per example}}                                           & 4+: .2\%         \\ \hline
\textbf{\begin{tabular}[c]{@{}l@{}}Number of examples w/ 2+ raters\\agreeing on at least 1 label\end{tabular}} & 54,263 (94\%)\\ \hline
\textbf{\begin{tabular}[c]{@{}l@{}}Number of examples w/ 3+ raters\\agreeing on at least 1 label\end{tabular}} & 17,763 (31\%) \\ \hline
\end{tabular}
    \end{centering}
    }
    \caption{Summary statistics of our labeled data.}
    \label{tab:data_stats}
\end{table}

Table~\ref{tab:data_stats} shows summary statistics for the data. Most of the examples (83\%) have a single emotion label and have at least two raters agreeing on a single label (94\%). The Neutral category makes up 26\% of all emotion labels -- we exclude that category from the following analyses, since we do not consider it to be part of the semantic space of emotions.

Figure~\ref{fig:emo_distr} shows the distribution of emotion labels. We can see a large disparity in terms of emotion frequencies (e.g. \emph{admiration} is 30 times more frequent than \emph{grief}), despite our emotion and sentiment balancing steps taken during data selection. This is expected given the disparate frequencies of emotions in natural human expression.

\subsection{Interrater Correlation}
\label{ssec:interrater_corr}

\begin{figure}[t!]
 \centering
   \centering
   \includegraphics[width=\linewidth]{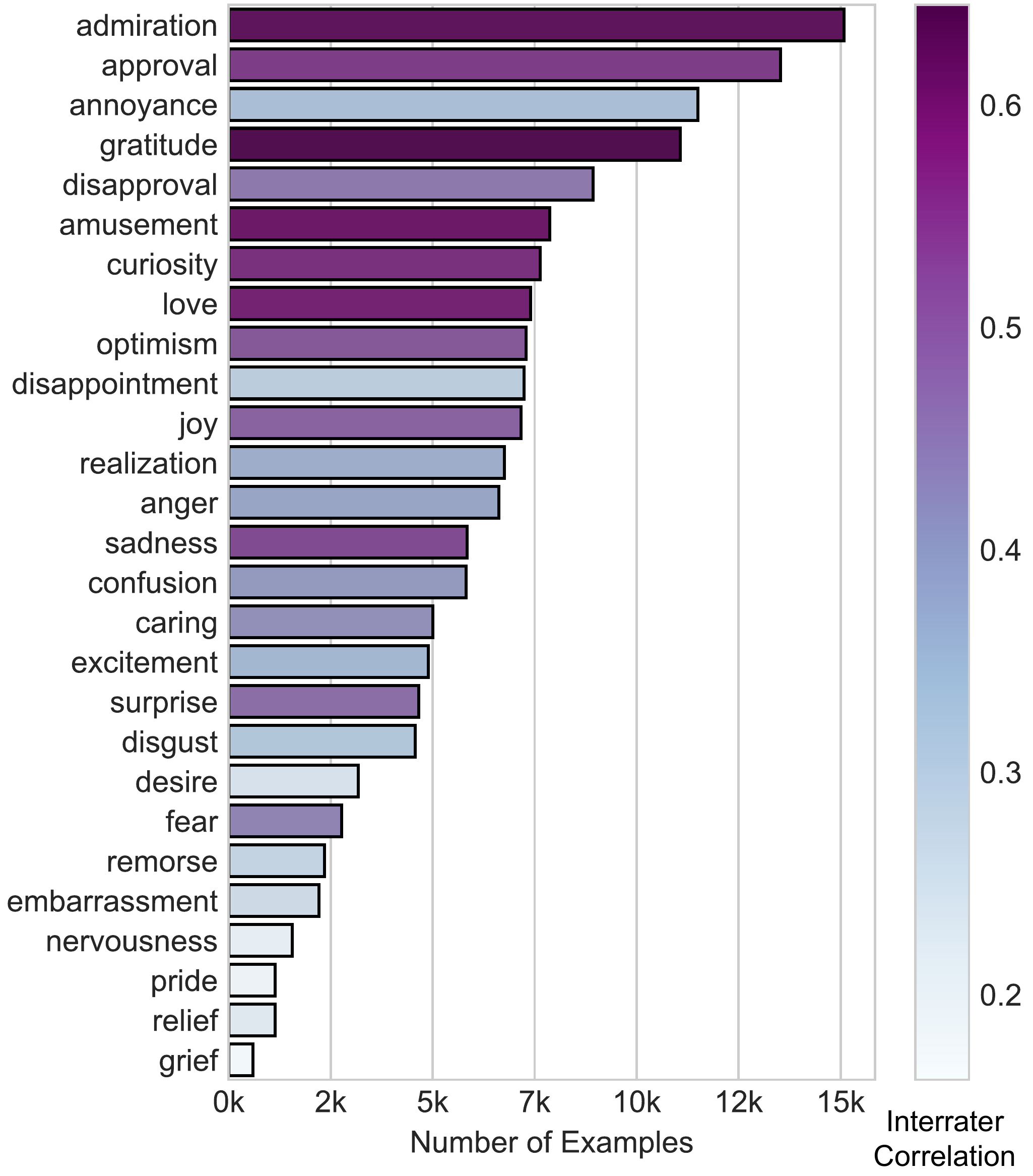}
   \caption{Our emotion categories, ordered by the number of examples where at least one rater uses a particular label. The color indicates the interrater correlation.}
   \label{fig:emo_distr}
\end{figure}

We estimate rater agreement for each emotion via interrater correlation \citep{delgado2019cohen}.\footnote{We use correlations as opposed to Cohen's kappa \cite{cohen1960coefficient} because the former is a more interpretable metric and it is also more suitable for measuring agreement among a variable number of raters rating different examples. In Appendix~\ref{sec:appendix_kappa} we report Cohen's kappa values as well, which correlate highly with the values obtained from interrater correlation (Pearson $r=0.85$, $p < 0.001$).} For each rater $r\in R$, we calculate the Spearman correlation between $r$'s judgments and the mean of other raters' judgments, for all examples that $r$ rated. We then take the average of these rater-level correlation scores. In Section~\ref{ssec:ppca}, we show that each emotion has significant interrater correlation, after controlling for several potential confounds.

Figure~\ref{fig:emo_distr} shows that \emph{gratitude}, \emph{admiration} and \emph{amusement} have the highest and \emph{grief} and \emph{nervousness} have the lowest interrater correlation. Emotion frequency correlates with interrater agreement but the two are not equivalent. Infrequent emotions can have relatively high interrater correlation (e.g., \emph{fear}), and frequent emotions can have have relatively low interrater correlation (e.g., \emph{annoyance}).

\subsection{Correlation Among Emotions}
\begin{figure}[t!]
\centering
   \includegraphics[width=1\linewidth]{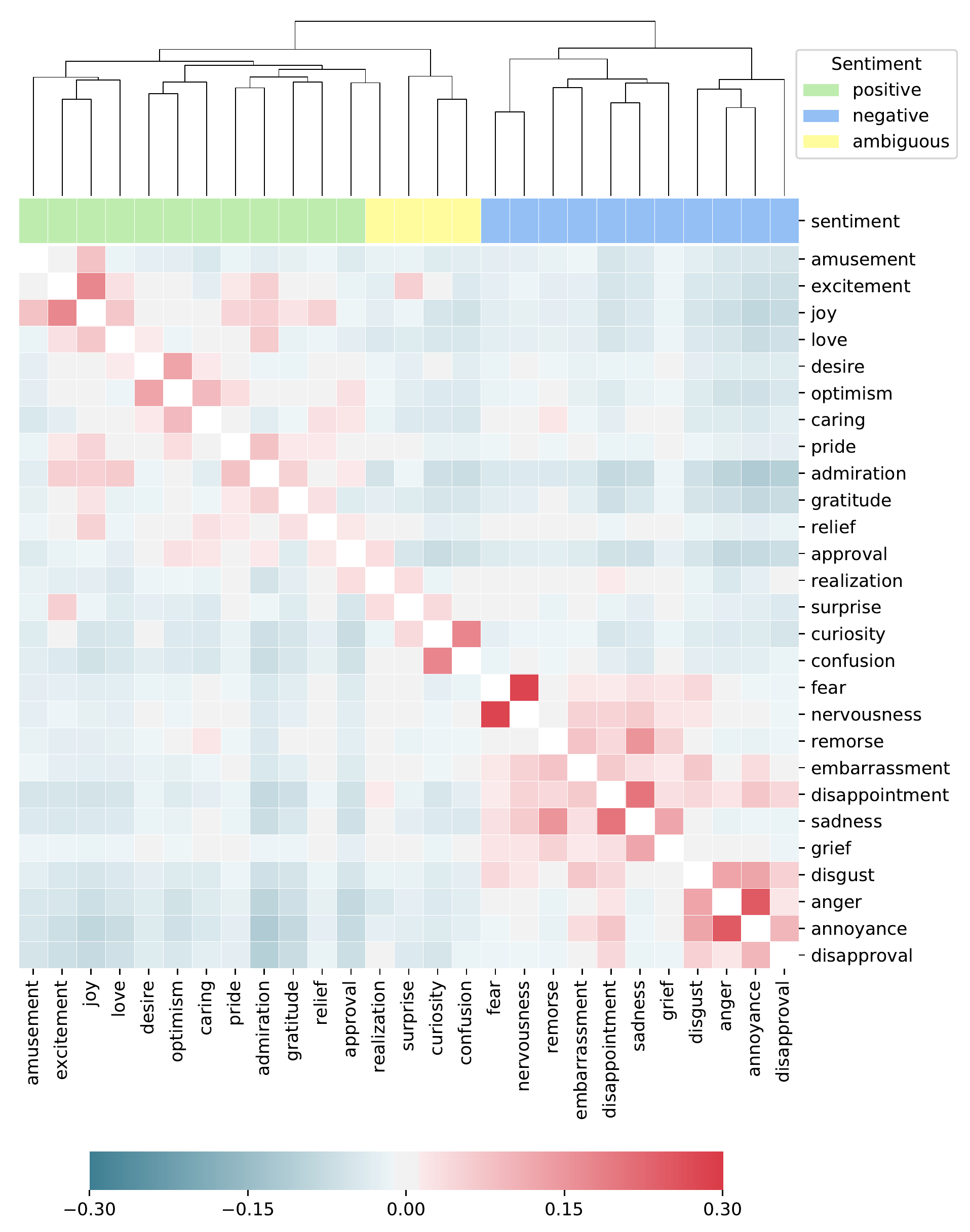}
   \caption{The heatmap shows the correlation between ratings for each emotion. The dendrogram represents the a hierarchical clustering of the ratings. The sentiment labeling was done \emph{a priori} and it shows that the clusters closely map onto sentiment groups. }
   \label{fig:hierarchical_corr}
\end{figure}

To better understand the relationship between emotions in our data, we look at their correlations. Let $N$ be the number of examples in our dataset. We obtain $N$ dimensional vectors for each emotion by averaging raters' judgments for all examples labeled with that emotion. We calculate Pearson correlation values between each pair of emotions.  The heatmap in Figure~\ref{fig:hierarchical_corr} shows that emotions that are related in intensity (e.g. \emph{annoyance} and \emph{anger}, \emph{joy} and \emph{excitement}, \emph{nervousness} and \emph{fear}) have a strong positive correlation. On the other hand, emotions that have the opposite sentiment are negatively correlated.

We also perform hierarchical clustering to uncover the nested structure of our taxonomy. We use correlation as a distance metric and ward as a linkage method, applied to the averaged ratings. The dendrogram on the top of  Figure~\ref{fig:hierarchical_corr} shows that emotions that are related by intensity are neighbors, and that larger clusters map closely onto sentiment categories. Interestingly, emotions that we labeled as ``ambiguous'' in terms of sentiment (e.g. \emph{surprise}) are closer to the positive than to the negative category. This suggests that in our data, ambiguous emotions are more likely to occur in the context of positive sentiment than that of negative sentiment.

\subsection{Principal Preserved Component Analysis}
\label{ssec:ppca}

\begin{algorithm}[]
\caption{Leave-One-Rater-Out PPCA}\label{alg:leaveout_ppca}
\begin{algorithmic}[1]
\STATE $R \leftarrow$ set of raters
\STATE $E \leftarrow$ set of emotions
\STATE $C \in \mathbb{R}^{|R| \times |E|}$
\FOR{all raters $r \in\{1,...,|R|\}$}
\STATE $n \leftarrow$ number of examples annotated by $r$ 
\STATE $J \in \mathbb{R}^{n\times|R|\times|E|} \leftarrow$ all ratings for the examples annotated by $r$
\STATE $J^{-r}\in \mathbb{R}^{n \times |R|-1 \times |E|} \leftarrow$ all ratings in $J$, excluding $r$ 
\STATE $J^{r}  \in \mathbb{R}^{n \times |E|} \leftarrow$ all ratings by $r$
\STATE $X, Y \in \mathbb{R}^{n \times |E|} \leftarrow$  randomly split $J^{-r}$ and average ratings across raters for both sets
\STATE $W \in \mathbb{R}^{|E| \times |E|} \leftarrow$  result of $PPCA(X, Y)$
\FOR{all components$^\dagger$ $\mathbf{w}_{i\in\{1,...,|E|\}}$ in $W$}
\STATE $\mathbf{v}^r_i \leftarrow$ projection$^\ddagger$ of $J^r$ onto $\mathbf{w}_i$
\STATE $\mathbf{v}_i^{-r} \leftarrow$ projection$^\ddagger$ of $J^{-r}$  onto $\mathbf{w}_i$
\STATE $C_{r,i} \leftarrow$ correlation between $\mathbf{v}^r_i$ and $\mathbf{v}_i^{-r}$, partialing out $\mathbf{v}^{-r}_k \forall k \in \{1,...,i-1\}$
\ENDFOR
\ENDFOR
\STATE $C' \leftarrow$ Wilcoxon signed rank test on $C$
\STATE $C'' \leftarrow$ Bonferroni correction on $C' (\alpha=0.05)$
\end{algorithmic}
$^\dagger$in descending order of eigenvalue\\
$^\ddagger$we demean vectors before projection
\end{algorithm}

\noindent To better understand agreement among raters and the latent structure of the emotion space, we apply Principal Preserved Component Analysis (PPCA) \citep{cowen2019primacy} to our data. PPCA extracts linear combinations of attributes (here, emotion judgments), that maximally covary across two sets of data that measure the same attributes (here, randomly split judgments for each example). Thus, PPCA allows us to uncover latent dimensions of emotion that have high agreement across raters.

Unlike Principal Component Analysis (PCA), PPCA examines the cross-covariance
between datasets rather than the variance–covariance matrix
within a single dataset. We obtain the principal preserved components (PPCs) of two datasets (matrices) $X, Y \in \mathbb{R}^{N \times |E|}$, where $N$ is the number of examples and $|E|$ is the number of emotions, by calculating the \emph{eigenvectors} of the symmetrized cross covariance matrix $X^TY + Y^TX$.

\paragraph{Extracting significant dimensions.} We remove examples labeled as Neutral, and keep those examples that still have at least 3 ratings after this filtering step. We then determine the number of significant dimensions using a leave-one-rater out analysis, as described by Algorithm~\ref{alg:leaveout_ppca}.

We find that all 27 PPCs are highly significant. Specifically, Bonferroni-corrected p-values are less than 1.5e-6 for all dimensions (corrected $\alpha$ = 0.0017), suggesting that the emotions were highly dissociable. Such a high degree of significance for all dimensions is nontrivial. For example, \citet{cowen2019primacy} find that only 12 out of their 30 emotion categories are significantly dissociable.

\paragraph{t-SNE projection.} To better understand how the examples are organized in the emotion space, we apply t-SNE, a dimension reduction method that seeks to preserve distances between data points, using the scikit-learn package \citep{scikit-learn}. The dataset can be explored in our interactive plot\footnote{\url{https://nlp.stanford.edu/~ddemszky/goemotions/tsne.html}}, where one can also look at the texts and the annotations. The color of each data point is the weighted average of the RGB values representing those emotions that at least half of the raters selected.

\subsection{Linguistic Correlates of Emotions}
\begin{table*}[t!]
    
    \resizebox{\linewidth}{!}{

    \begin{centering}
\begin{tabular}{|cccc|ccccc|} \hline
\cellcolor[HTML]{BEECAF}\textbf{admiration} & \cellcolor[HTML]{BEECAF}\textbf{amusement}  & \cellcolor[HTML]{BEECAF}\textbf{approval}  & \cellcolor[HTML]{BEECAF}\textbf{caring} & \cellcolor[HTML]{A6CBF7}\textbf{anger}       & \cellcolor[HTML]{A6CBF7}\textbf{annoyance}     & \cellcolor[HTML]{A6CBF7}\textbf{disappointment} & \multicolumn{1}{c|}{\textbf{\cellcolor[HTML]{A6CBF7}\textbf{disapproval}}}  & \cellcolor[HTML]{FFFC9E}\textbf{confusion} \\
great (42)                               & lol (66)                                 & agree (24)                              & you (12)                             & fuck (24)                                 & annoying (14)                               & disappointing (11)                           & \multicolumn{1}{c|}{not (16)}                                   & confused (18)                           \\
awesome (32)                             & haha (32)                                & not (13)                                & worry (11)                           & hate (18)                                 & stupid (13)                                 & disappointed (10)                            & \multicolumn{1}{c|}{don't (14)}                                 & why (11)                                \\
amazing (30)                             & funny (27)                               & don't (12)                              & careful (9)                          & fucking (18)                              & fucking (12)                                & bad (9)                                      & \multicolumn{1}{c|}{disagree (9)}                              & sure (10)                               \\
good (28)                                & lmao (21)                                & yes (12)                                & stay (9)                             & angry (11)                                & shit (10)                                    & disappointment (7)                           & \multicolumn{1}{c|}{nope (8)}                                  & what (10)                                \\
beautiful (23)                           & hilarious (18)                           & agreed (11)                             & your (8)                             & dare (10)                                 & dumb (9)                                    & unfortunately (7)                            & \multicolumn{1}{c|}{doesn't (7)}                               & understand (8)                        \\ \hline
\cellcolor[HTML]{BEECAF}\textbf{desire}     & \cellcolor[HTML]{BEECAF}\textbf{excitement} & \cellcolor[HTML]{BEECAF}\textbf{gratitude} & \cellcolor[HTML]{BEECAF}\textbf{joy}    & \cellcolor[HTML]{A6CBF7}\textbf{disgust}     & \cellcolor[HTML]{A6CBF7}\textbf{embarrassment} & \cellcolor[HTML]{A6CBF7}\textbf{fear}           & \multicolumn{1}{c|}{\cellcolor[HTML]{A6CBF7}\textbf{grief} }      & \cellcolor[HTML]{FFFC9E}\textbf{curiosity} \\
wish (29)                                & excited (21)                             & thanks (75)                             & happy (32)                           & disgusting (22)                           & embarrassing (12)                           & scared (16)                                  & \multicolumn{1}{c|}{died (6) }                                 & curious (22)                            \\
want (8)                                 & happy (8)                                & thank (69)                              & glad (27)                            & awful (14)                                & shame (11)                                  & afraid (16)                                  & \multicolumn{1}{c|}{rip (4)}                                   & what (18)                               \\
wanted (6)                               & cake (8)                                 & for (24)                                & enjoy (20)                           & worst (13)                                & awkward (10)                                & scary (15)                                   &\multicolumn{1}{c|}{}  & why (13)                                \\
could (6)                                & wow (8)                                  & you (18)                                & enjoyed (12)                         & worse (12)                                & embarrassment (8)                           & terrible (12)                                &      \multicolumn{1}{c|}{}                          & how (11)                                \\
ambitious (4)                            & interesting (7)                          & sharing (17)                            & fun (12)                             & weird (9)                                 & embarrassed (7)                             &  terrifying (11)                              &   \multicolumn{1}{c|}{}                                            & did (10)                                \\ \hline
\cellcolor[HTML]{BEECAF}\textbf{love}       & \cellcolor[HTML]{BEECAF}\textbf{optimism}   & \cellcolor[HTML]{BEECAF}\textbf{pride}     & \cellcolor[HTML]{BEECAF}\textbf{relief} & \cellcolor[HTML]{A6CBF7}\textbf{nervousness} & \cellcolor[HTML]{A6CBF7}\textbf{remorse}       & \multicolumn{1}{c|}{\cellcolor[HTML]{A6CBF7}\textbf{sadness}}        & \cellcolor[HTML]{FFFC9E}\textbf{realization}  & \cellcolor[HTML]{FFFC9E}\textbf{surprise}  \\
love (76)                                & hope (45)                                & proud (14)                              & glad (5)                             & nervous (8)                               & sorry (39)                                  & \multicolumn{1}{c|}{sad (31) }                                    & realize (14)                               & wow (23)                                \\
loved (21)                               & hopefully (19)                           & pride (4)                               & relieved (4)                         & worried (8)                               & regret (9)                                  & \multicolumn{1}{c|}{sadly (16)}                                   & realized (12)                              & surprised (21)                          \\
favorite (13)                            & luck (18)                                & accomplishment                     & relieving (4)                        & anxiety (6)                               & apologies (7)                               & \multicolumn{1}{c|}{sorry (15)}                                   & realised (7)                               & wonder (15)                             \\
loves (12)                               & hoping (16)                              &                       (4)               & relief (4)                           & anxious (4)                               & apologize (6)                               & \multicolumn{1}{c|}{painful (10)    }                             & realization (6)                            & shocked (12)                            \\
like (9)                                 & will (8)                                 &                                            &                                         & worrying (4)                              & guilt (5)                                   & \multicolumn{1}{c|}{crying (9)}                                   & thought (6)                                & omg (11)                                \\
\hline
\end{tabular}
    \end{centering}
    }
    \caption{Top 5 words associated with each emotion (\colorbox[HTML]{BEECAF}{positive},\colorbox[HTML]{A6CBF7}{negative}, \colorbox[HTML]{FFFC9E}{ambiguous}). The rounded $z$-scored log odds ratios in the parentheses, with the threshold set at 3, indicate significance of association.}
    \label{tab:emo_words}
\end{table*}

We extract the lexical correlates of each emotion by calculating the log odds ratio, informative Dirichlet prior \citep{monroe2008fightin} of all tokens for each emotion category contrasting to all other emotions. Since the log odds are $z$-scored, all values greater than 3 indicate highly significant ($>$3 std) association with the corresponding emotion. We list the top 5 tokens for each category in Table~\ref{tab:emo_words}. We find that those emotions that are highly significantly associated with certain tokens (e.g. \emph{gratitude} with ``thanks'', \emph{amusement} with ``lol'') tend to have the highest interrater correlation (see Figure~\ref{fig:emo_distr}). Conversely, emotions that have fewer significantly associated tokens (e.g. \emph{grief} and \emph{nervousness}) tend to have low interrater correlation. These results suggest certain emotions are more verbally implicit and may require more context to be interpreted.
\section{Modeling}

We present a strong baseline emotion prediction model for GoEmotions.

\subsection{Data Preparation}

To minimize the noise in our data, we filter out emotion labels selected by only a single annotator. We keep examples with at least one label after this filtering is performed --- this amounts to 93\% of the original data. We randomly split this data into train (80\%), dev (10\%) and test (10\%) sets. We only evaluate on the test set once the model is finalized.

Even though we filter our data for the baseline experiments, we see particular value in the 4K examples that lack agreement. This subset of the data likely contains edge/difficult examples for the emotion domain (e.g., emotion-ambiguous text), and present challenges for further exploration. That is why we release all 58K examples with all annotators' ratings.

\paragraph{Grouping emotions.} We create a hierarchical grouping of our taxonomy, and evaluate the model performance on each level of the hierarchy. A sentiment level divides the labels into 4 categories -- \emph{positive}, \emph{negative}, \emph{ambiguous} and Neutral -- with the Neutral category intact, and the rest of the mapping as shown in Figure~\ref{fig:hierarchical_corr}. The Ekman level further divides the taxonomy using the Neutral label and the following 6 groups: \emph{anger} (maps to: \emph{anger}, \emph{annoyance}, \emph{disapproval}), \emph{disgust} (maps to: disgust), \emph{fear} (maps to: \emph{fear}, \emph{nervousness}), \emph{joy} (all \emph{positive} emotions), \emph{sadness} (maps to: \emph{sadness}, \emph{disappointment}, \emph{embarrassment}, \emph{grief}, \emph{remorse}) and \emph{surprise} (all \emph{ambiguous} emotions).


\subsection{Model Architecture}

We use the BERT-base model \cite{devlin2019bert} for our experiments. We add a dense output layer on top of the pretrained model for the purposes of finetuning, with a sigmoid cross entropy loss function to support multi-label classification. As an additional baseline, we train a bidirectional LSTM.

\subsection{Parameter Settings}

When finetuning the pre-trained BERT model, we keep most of the hyperparameters set by \citet{devlin2019bert} intact and only change the batch size and learning rate. We find that training for at least 4 epochs is necessary for learning the data, but training for more epochs results in overfitting. We also find that a small batch size of 16 and learning rate of 5e-5 yields the best performance.

For the biLSTM, we set the hidden layer dimensionality to 256, the learning rate to 0.1, with a decay rate of 0.95. We apply a dropout of 0.7.

\subsection{Results}

Table~\ref{tab:results} summarizes the performance of our best model, BERT, on the test set, which achieves an average F1-score of .46 (std=.19). The model obtains the best performance on emotions with overt lexical markers, such as \emph{gratitude} (.86), \emph{amusement} (.8) and \emph{love} (.78). The model obtains the lowest F1-score on \emph{grief} (0), \emph{relief} (.15) and \emph{realization} (.21), which are the lowest frequency emotions. We find that less frequent emotions tend to be confused by the model with more frequent emotions related in sentiment and intensity (e.g., \emph{grief} with \emph{sadness}, \emph{pride} with \emph{admiration}, \emph{nervousness} with \emph{fear}) --- see Appendix~\ref{sec:appendix_confusion_matrix} for a more detailed analysis.

Table~\ref{tab:sentiment_results} and Table~\ref{tab:ekman_results} show results for a sentiment-grouped model (F1-score = .69) and an Ekman-grouped model (F1-score = .64), respectively.
The significant performance increase in the transition from full to Ekman-level taxonomy indicates that this grouping mitigates confusion among inner-group lower-level categories.

The biLSTM model performs significantly worse than BERT, obtaining an average F1-score of .41 for the full taxonomy, .53 for an Ekman-grouped model and .6 for a sentiment-grouped model.

\begin{table}[h!]
    \centering
    \resizebox{.95\linewidth}{!}{
    \begin{centering}
\begin{tabular}{@{}lccc@{}}
\toprule
\textbf{Emotion}       & \textbf{Precision} & \textbf{Recall} & \textbf{F1}   \\ \midrule
admiration             & 0.53               & 0.83            & 0.65          \\
amusement              & 0.70               & 0.94            & 0.80          \\
anger                  & 0.36               & 0.66            & 0.47          \\
annoyance              & 0.24               & 0.63            & 0.34          \\
approval               & 0.26               & 0.57            & 0.36          \\
caring                 & 0.30               & 0.56            & 0.39          \\
confusion              & 0.24               & 0.76            & 0.37          \\
curiosity              & 0.40               & 0.84            & 0.54          \\
desire                 & 0.43               & 0.59            & 0.49          \\
disappointment         & 0.19               & 0.52            & 0.28          \\
disapproval            & 0.29               & 0.61            & 0.39          \\
disgust                & 0.34               & 0.66            & 0.45          \\
embarrassment          & 0.39               & 0.49            & 0.43          \\
excitement             & 0.26               & 0.52            & 0.34          \\
fear                   & 0.46               & 0.85            & 0.60          \\
gratitude              & 0.79               & 0.95            & 0.86          \\
grief                  & 0.00               & 0.00            & 0.00          \\
joy                    & 0.39               & 0.73            & 0.51          \\
love                   & 0.68               & 0.92            & 0.78          \\
nervousness            & 0.28               & 0.48            & 0.35          \\
neutral                & 0.56               & 0.84            & 0.68          \\
optimism               & 0.41               & 0.69            & 0.51          \\
pride                  & 0.67               & 0.25            & 0.36          \\
realization            & 0.16               & 0.29            & 0.21          \\
relief                 & 0.50               & 0.09            & 0.15          \\
remorse                & 0.53               & 0.88            & 0.66          \\
sadness                & 0.38               & 0.71            & 0.49          \\
surprise               & 0.40               & 0.66            & 0.50          \\
\textbf{macro-average} & \textbf{0.40}      & \textbf{0.63}   & \textbf{0.46} \\
\textbf{std}     & \textbf{0.18}      & \textbf{0.24}   & \textbf{0.19} \\
\bottomrule
\end{tabular}
    \end{centering}
    }
    \caption{Results based on GoEmotions taxonomy.}
    \label{tab:results}
\end{table}

\begin{table}[h!]
\centering
    
    \resizebox{.9\linewidth}{!}{

    \begin{centering}
\begin{tabular}{@{}lccc@{}}
\toprule
\textbf{Sentiment}     & \textbf{Precision} & \textbf{Recall} & \textbf{F1}   \\ \midrule
ambiguous              & 0.54               & 0.66            & 0.60          \\
negative               & 0.65               & 0.76            & 0.70          \\
neutral                & 0.64               & 0.69            & 0.67          \\
positive               & 0.78               & 0.87            & 0.82          \\
\textbf{macro-average} & \textbf{0.65}      & \textbf{0.74}   & \textbf{0.69} \\
\textbf{std}         & \textbf{0.09}      & \textbf{0.10}   & \textbf{0.09} \\ \bottomrule
\end{tabular}
    \end{centering}
    }
    \caption{Results based on sentiment-grouped data.}
    \label{tab:sentiment_results}
\end{table}

\begin{table}[h!]
\centering
    
    \resizebox{.95\linewidth}{!}{

    \begin{centering}
\begin{tabular}{@{}lccc@{}}
\toprule
\textbf{Ekman Emotion} & \textbf{Precision} & \textbf{Recall} & \textbf{F1}   \\ \midrule
anger                  & 0.50               & 0.65            & 0.57          \\
disgust                & 0.52               & 0.53            & 0.53          \\
fear                   & 0.61               & 0.76            & 0.68          \\
joy                    & 0.77               & 0.88            & 0.82          \\
neutral                & 0.66               & 0.67            & 0.66          \\
sadness                & 0.56               & 0.62            & 0.59          \\
surprise                & 0.53              & 0.70            & 0.61          \\
\textbf{macro-average} & \textbf{0.59}      & \textbf{0.69}   & \textbf{0.64} \\
\textbf{std}         & \textbf{0.10}      & \textbf{0.11}   & \textbf{0.10} \\ \bottomrule
\end{tabular}
    \end{centering}
    }
    \caption{Results using Ekman's taxonomy.}
    \label{tab:ekman_results}
\end{table}

\section{Transfer Learning Experiments}

We conduct transfer learning experiments on existing emotion benchmarks, in order to show our data generalizes across domains and taxonomies. The goal is to demonstrate that given little labeled data in a target domain, one can utilize GoEmotions as baseline emotion understanding data.

\begin{figure*}[t!]
 \centering
   \centering
   \includegraphics[width=\linewidth]{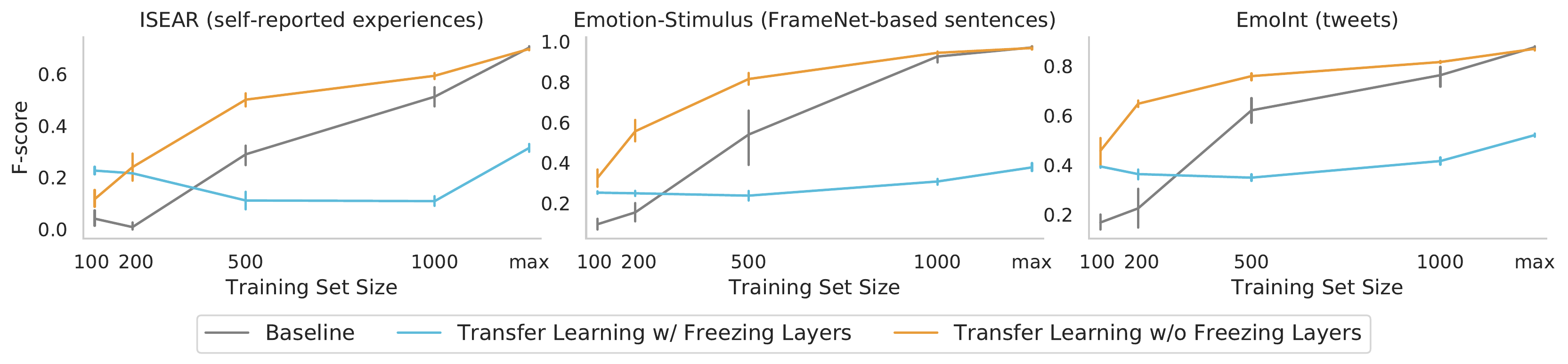}
   \caption{Transfer learning results in terms of average F1-scores across emotion categories. The bars indicate the 95\% confidence intervals, which we obtain from 10 different runs on 10 different random splits of the data.}
   \label{fig:transfer_results}
\end{figure*}

\subsection{Emotion Benchmark Datasets}
We consider the nine benchmark datasets from \citet{unified-klinger2018analysis}'s Unified Dataset, which vary in terms of their size, domain, quality and taxonomy. In the interest of space, we only discuss three of these datasets here, chosen based on their diversity of domains. In our experiments, we observe similar trends for the additional benchmarks, and all are included in the Appendix~\ref{sec:appendix_transfer_learning}.

The International Survey on Emotion Antecedents and Reactions (ISEAR) \citep{scherer1994evidence} is a collection of personal reports on emotional events, written by 3000 people from different cultural backgrounds. The dataset contains 8k sentences, each labeled with a single emotion. The categories are \emph{anger}, \emph{disgust}, \emph{fear}, \emph{guilt}, \emph{joy}, \emph{sadness} and \emph{shame}.


EmoInt \citep{mohammad-etal-2018-semeval} is part of the SemEval 2018 benchmark, and it contains crowdsourced annotations for 7k tweets. The labels are intensity annotations for \emph{anger}, \emph{joy}, \emph{sadness}, and \emph{fear}. We obtain binary annotations for these emotions by using .5 as the cutoff.

Emotion-Stimulus \citep{ghazi2015detecting} contains annotations for 2.4k sentences generated based on FrameNet's emotion-directed frames. Their taxonomy is \emph{anger}, \emph{disgust}, \emph{fear}, \emph{joy}, \emph{sadness}, \emph{shame} and \emph{surprise}.

\subsection{Experimental Setup}
\label{ssec:transfer_learning_setup}

\paragraph{Training set size.} We experiment with varying amount of training data from the target domain dataset, including 100, 200, 500, 1000, and 80\% (named ``max'') of dataset examples. We generate 10 random splits for each train set size, with the remaining examples held as a test set.

We report the results of the finetuning experiments detailed below for each data size, with confidence intervals based on repeated experiments using the splits.


\paragraph{Finetuning.} We compare three different finetuning setups. In the \textsc{baseline} setup, we finetune BERT only on the target dataset. In the \textsc{freeze} setup, we first finetune BERT on GoEmotions, then perform transfer learning by replacing the final dense layer, freezing all layers besides the last layer and finetuning on the target dataset. The \textsc{nofreeze} setup is the same as \textsc{freeze}, except that we do not freeze the bottom layers. We hold the batch size at 16, learning rate at 2e-5 and number of epochs at 3 for all experiments.

\subsection{Results}

The results in Figure~\ref{fig:transfer_results} suggest that our dataset generalizes well to different domains and taxonomies, and that using a model using GoEmotions can help in cases when there is limited data from the target domain, or limited resources for labeling.
 
 Given limited target domain data (100 or 200 examples), both \textsc{freeze} and \textsc{nofreeze} yield significantly higher performance than the \textsc{baseline}, for all three datasets. Importantly, \textsc{nofreeze} results show significantly higher performance for all training set sizes, except for ``max'', where \textsc{nofreeze} and \textsc{baseline} perform similarly.


\section{Conclusion}
We present GoEmotions, a large, manually annotated, carefully curated dataset for fine-grained emotion prediction. We provide a detailed data analysis, demonstrating the reliability of the annotations for the full taxonomy. We show the generalizability of the data across domains and taxonomies via transfer learning experiments. We build a strong baseline by fine-tuning a BERT model, however, the results suggest much room for future improvement.
Future work can explore the cross-cultural robustness of emotion ratings, and extend the taxonomy to other languages and domains.

\textbf{Data Disclaimer:} We are aware that the dataset contains biases and is not representative of global diversity. We are aware that the dataset contains potentially problematic content. Potential biases in the data include: Inherent biases in Reddit and user base biases, the offensive/vulgar word lists used for data filtering, inherent or unconscious bias in assessment of offensive identity labels, annotators were all native English speakers from India. All these likely affect labeling, precision, and recall for a trained model.
The emotion pilot model used for sentiment labeling, was trained on examples reviewed by the research team. Anyone using this dataset should be aware of these limitations of the dataset.



\section*{Acknowledgments}
We thank the three anonymous reviewers for their constructive feedback. We would also like to thank the annotators for their hard work.

\bibliography{acl2020}

\begin{thebibliography}{42}
\expandafter\ifx\csname natexlab\endcsname\relax\def\natexlab#1{#1}\fi

\bibitem[{Abdul-Mageed and Ungar(2017)}]{abdul2017emonet}
Muhammad Abdul-Mageed and Lyle Ungar. 2017.
\newblock {EmoNet: Fine-Grained Emotion Detection with Gated Recurrent Neural
  Networks}.
\newblock In \emph{Proceedings of the 55th Annual Meeting of the Association
  for Computational Linguistics (Volume 1: Long Papers)}, pages 718--728.

\bibitem[{Alm et~al.(2005)Alm, Roth, and Sproat}]{alm-etal-2005-emotions}
Cecilia~Ovesdotter Alm, Dan Roth, and Richard Sproat. 2005.
\newblock \href {https://www.aclweb.org/anthology/H05-1073} {Emotions from
  text: Machine learning for text-based emotion prediction}.
\newblock In \emph{Proceedings of Human Language Technology Conference and
  Conference on Empirical Methods in Natural Language Processing}, pages
  579--586, Vancouver, British Columbia, Canada. Association for Computational
  Linguistics.

\bibitem[{Bostan and Klinger(2018)}]{unified-klinger2018analysis}
Laura-Ana-Maria Bostan and Roman Klinger. 2018.
\newblock An analysis of annotated corpora for emotion classification in text.
\newblock In \emph{Proceedings of the 27th International Conference on
  Computational Linguistics}, pages 2104--2119.

\bibitem[{Breitfeller et~al.(2019)Breitfeller, Ahn, Jurgens, and
  Tsvetkov}]{breitfeller2019finding}
Luke Breitfeller, Emily Ahn, David Jurgens, and Yulia Tsvetkov. 2019.
\newblock Finding microaggressions in the wild: A case for locating elusive
  phenomena in social media posts.
\newblock In \emph{Proceedings of the 2019 Conference on Empirical Methods in
  Natural Language Processing and the 9th International Joint Conference on
  Natural Language Processing (EMNLP-IJCNLP)}, pages 1664--1674.

\bibitem[{Buechel and Hahn(2017)}]{buechel2017emobank}
Sven Buechel and Udo Hahn. 2017.
\newblock {Emobank: Studying the impact of annotation perspective and
  representation format on dimensional emotion analysis}.
\newblock In \emph{Proceedings of the 15th Conference of the European Chapter
  of the Association for Computational Linguistics: Volume 2, Short Papers},
  pages 578--585.

\bibitem[{Cohen(1960)}]{cohen1960coefficient}
Jacob Cohen. 1960.
\newblock A coefficient of agreement for nominal scales.
\newblock \emph{Educational and psychological measurement}, 20(1):37--46.

\bibitem[{Cowen et~al.(2019{\natexlab{a}})Cowen, Sauter, Tracy, and
  Keltner}]{cowen2019mapping}
Alan Cowen, Disa Sauter, Jessica~L Tracy, and Dacher Keltner.
  2019{\natexlab{a}}.
\newblock Mapping the passions: Toward a high-dimensional taxonomy of emotional
  experience and expression.
\newblock \emph{Psychological Science in the Public Interest}, 20(1):69--90.

\bibitem[{Cowen et~al.(2018)Cowen, Elfenbein, Laukka, and
  Keltner}]{cowen2018mapping}
Alan~S Cowen, Hillary~Anger Elfenbein, Petri Laukka, and Dacher Keltner. 2018.
\newblock Mapping 24 emotions conveyed by brief human vocalization.
\newblock \emph{American Psychologist}, 74(6):698--712.

\bibitem[{Cowen et~al.(in press)Cowen, Fang, Sauter, and
  Keltner}]{cowen2019music}
Alan~S Cowen, Xia Fang, Disa Sauter, and Dacher Keltner. in press.
\newblock What music makes us feel: At least thirteen dimensions organize
  subjective experiences associated with music across cultures.
\newblock \emph{Proceedings of the National Academy of Sciences}.

\bibitem[{Cowen and Keltner(2017)}]{cowen2017self}
Alan~S Cowen and Dacher Keltner. 2017.
\newblock Self-report captures 27 distinct categories of emotion bridged by
  continuous gradients.
\newblock \emph{Proceedings of the National Academy of Sciences},
  114(38):E7900--E7909.

\bibitem[{Cowen and Keltner(2019)}]{cowen2019face}
Alan~S Cowen and Dacher Keltner. 2019.
\newblock What the face displays: Mapping 28 emotions conveyed by naturalistic
  expression.
\newblock \emph{American Psychologist}.

\bibitem[{Cowen et~al.(2019{\natexlab{b}})Cowen, Laukka, Elfenbein, Liu, and
  Keltner}]{cowen2019primacy}
Alan~S Cowen, Petri Laukka, Hillary~Anger Elfenbein, Runjing Liu, and Dacher
  Keltner. 2019{\natexlab{b}}.
\newblock The primacy of categories in the recognition of 12 emotions in speech
  prosody across two cultures.
\newblock \emph{Nature Human Behaviour}, 3(4):369.

\bibitem[{CrowdFlower(2016)}]{crowdflower-2016}
CrowdFlower. 2016.
\newblock \href
  {https://www.figure-eight.com/data/sentiment-analysis-emotion-text/}
  {https://www.figure-eight.com/data/sentiment-analysis-emotion-text/}.

\bibitem[{Delgado and Tibau(2019)}]{delgado2019cohen}
Rosario Delgado and Xavier-Andoni Tibau. 2019.
\newblock Why cohen’s kappa should be avoided as performance measure in
  classification.
\newblock \emph{PloS one}, 14(9).

\bibitem[{Devlin et~al.(2019)Devlin, Chang, Lee, and
  Toutanova}]{devlin2019bert}
Jacob Devlin, Ming-Wei Chang, Kenton Lee, and Kristina Toutanova. 2019.
\newblock Bert: Pre-training of deep bidirectional transformers for language
  understanding.
\newblock In \emph{{17th Annual Conference of the North American Chapter of the
  Association for Computational Linguistics (NAACL)}}.

\bibitem[{Duggan and Smith(2013)}]{duggan20136}
Maeve Duggan and Aaron Smith. 2013.
\newblock 6\% of online adults are reddit users.
\newblock \emph{Pew Internet \& American Life Project}, 3:1--10.

\bibitem[{Ekman(1992{\natexlab{a}})}]{ekman1992there}
Paul Ekman. 1992{\natexlab{a}}.
\newblock \href {https://doi.org/10.1037/0033-295x.99.3.550} {Are there basic
  emotions?}
\newblock \emph{Psychological Review}, 99(3):550--553.

\bibitem[{Ekman(1992{\natexlab{b}})}]{ekman1992argument}
Paul Ekman. 1992{\natexlab{b}}.
\newblock An argument for basic emotions.
\newblock \emph{{Cognition \& Emotion}}, 6(3-4):169--200.

\bibitem[{Ghazi et~al.(2015)Ghazi, Inkpen, and Szpakowicz}]{ghazi2015detecting}
Diman Ghazi, Diana Inkpen, and Stan Szpakowicz. 2015.
\newblock {Detecting Emotion Stimuli in Emotion-Bearing Sentences}.
\newblock In \emph{International Conference on Intelligent Text Processing and
  Computational Linguistics}, pages 152--165. Springer.

\bibitem[{Hsu and Ku(2018)}]{hsu-ku-2018-socialnlp}
Chao-Chun Hsu and Lun-Wei Ku. 2018.
\newblock \href {https://doi.org/10.18653/v1/W18-3505} {{S}ocial{NLP} 2018
  {E}motion{X} challenge overview: Recognizing emotions in dialogues}.
\newblock In \emph{Proceedings of the Sixth International Workshop on Natural
  Language Processing for Social Media}, pages 27--31, Melbourne, Australia.
  Association for Computational Linguistics.

\bibitem[{Li et~al.(2017)Li, Su, Shen, Li, Cao, and Niu}]{li2017dailydialog}
Yanran Li, Hui Su, Xiaoyu Shen, Wenjie Li, Ziqiang Cao, and Shuzi Niu. 2017.
\newblock Dailydialog: A manually labelled multi-turn dialogue dataset.
\newblock \emph{arXiv preprint arXiv:1710.03957}.

\bibitem[{Liu et~al.(2019)Liu, Osama, and De~Andrade}]{liu2019dens}
Chen Liu, Muhammad Osama, and Anderson De~Andrade. 2019.
\newblock Dens: A dataset for multi-class emotion analysis.
\newblock \emph{arXiv preprint arXiv:1910.11769}.

\bibitem[{Mohammad(2018)}]{mohammad2018obtaining}
Saif Mohammad. 2018.
\newblock Obtaining reliable human ratings of valence, arousal, and dominance
  for 20,000 english words.
\newblock In \emph{Proceedings of the 56th Annual Meeting of the Association
  for Computational Linguistics (Volume 1: Long Papers)}, pages 174--184.

\bibitem[{Mohammad et~al.(2018)Mohammad, Bravo-Marquez, Salameh, and
  Kiritchenko}]{mohammad-etal-2018-semeval}
Saif Mohammad, Felipe Bravo-Marquez, Mohammad Salameh, and Svetlana
  Kiritchenko. 2018.
\newblock \href {https://doi.org/10.18653/v1/S18-1001} {{S}em{E}val-2018 task
  1: Affect in tweets}.
\newblock In \emph{Proceedings of The 12th International Workshop on Semantic
  Evaluation}, pages 1--17, New Orleans, Louisiana. Association for
  Computational Linguistics.

\bibitem[{Mohammad(2012)}]{tec-mohammad2012emotional}
Saif~M Mohammad. 2012.
\newblock \# emotional tweets.
\newblock In \emph{Proceedings of the First Joint Conference on Lexical and
  Computational Semantics-Volume 1: Proceedings of the main conference and the
  shared task, and Volume 2: Proceedings of the Sixth International Workshop on
  Semantic Evaluation}, pages 246--255. Association for Computational
  Linguistics.

\bibitem[{Mohammad et~al.(2015)Mohammad, Zhu, Kiritchenko, and
  Martin}]{electoraltweets-mohammad2015sentiment}
Saif~M Mohammad, Xiaodan Zhu, Svetlana Kiritchenko, and Joel Martin. 2015.
\newblock Sentiment, emotion, purpose, and style in electoral tweets.
\newblock \emph{Information Processing \& Management}, 51(4):480--499.

\bibitem[{Mohan et~al.(2017)Mohan, Guha, Harris, Popowich, Schuster, and
  Priebe}]{mohan2017impact}
Shruthi Mohan, Apala Guha, Michael Harris, Fred Popowich, Ashley Schuster, and
  Chris Priebe. 2017.
\newblock The impact of toxic language on the health of reddit communities.
\newblock In \emph{Canadian Conference on Artificial Intelligence}, pages
  51--56. Springer.

\bibitem[{Monroe et~al.(2008)Monroe, Colaresi, and Quinn}]{monroe2008fightin}
Burt~L Monroe, Michael~P Colaresi, and Kevin~M Quinn. 2008.
\newblock {Fightin'words: Lexical feature selection and evaluation for
  identifying the content of political conflict}.
\newblock \emph{Political Analysis}, 16(4):372--403.

\bibitem[{{\"O}hman et~al.(2018){\"O}hman, Kajava, Tiedemann, and
  Honkela}]{ohman2018creating}
Emily {\"O}hman, Kaisla Kajava, J{\"o}rg Tiedemann, and Timo Honkela. 2018.
\newblock Creating a dataset for multilingual fine-grained emotion-detection
  using gamification-based annotation.
\newblock In \emph{Proceedings of the 9th Workshop on Computational Approaches
  to Subjectivity, Sentiment and Social Media Analysis}, pages 24--30.

\bibitem[{Pedregosa et~al.(2011)Pedregosa, Varoquaux, Gramfort, Michel,
  Thirion, Grisel, Blondel, Prettenhofer, Weiss, Dubourg, Vanderplas, Passos,
  Cournapeau, Brucher, Perrot, and Duchesnay}]{scikit-learn}
F.~Pedregosa, G.~Varoquaux, A.~Gramfort, V.~Michel, B.~Thirion, O.~Grisel,
  M.~Blondel, P.~Prettenhofer, R.~Weiss, V.~Dubourg, J.~Vanderplas, A.~Passos,
  D.~Cournapeau, M.~Brucher, M.~Perrot, and E.~Duchesnay. 2011.
\newblock Scikit-learn: Machine learning in {P}ython.
\newblock \emph{Journal of Machine Learning Research}, 12:2825--2830.

\bibitem[{Peters et~al.(2018)Peters, Neumann, Iyyer, Gardner, Clark, Lee, and
  Zettlemoyer}]{peters2018deep}
Matthew~E. Peters, Mark Neumann, Mohit Iyyer, Matt Gardner, Christopher Clark,
  Kenton Lee, and Luke Zettlemoyer. 2018.
\newblock {Deep Contextualized Word Representations}.
\newblock In \emph{{16th Annual Conference of the North American Chapter of the
  Association for Computational Linguistics (NAACL)}}.

\bibitem[{Picard(1997)}]{picard1997affective}
Rosalind~W Picard. 1997.
\newblock \emph{{Affective Computing}}.
\newblock MIT Press.

\bibitem[{Pirina and {\c{C}}{\"o}ltekin(2018)}]{pirina2018identifying}
Inna Pirina and {\c{C}}a{\u{g}}r{\i} {\c{C}}{\"o}ltekin. 2018.
\newblock Identifying depression on reddit: The effect of training data.
\newblock In \emph{Proceedings of the 2018 EMNLP Workshop SMM4H: The 3rd Social
  Media Mining for Health Applications Workshop \& Shared Task}, pages 9--12.

\bibitem[{Plutchik(1980)}]{plutchik1980general}
Robert Plutchik. 1980.
\newblock A general psychoevolutionary theory of emotion.
\newblock In \emph{Theories of emotion}, pages 3--33. Elsevier.

\bibitem[{Russell(2003)}]{russell2003core}
James~A Russell. 2003.
\newblock Core affect and the psychological construction of emotion.
\newblock \emph{Psychological review}, 110(1):145.

\bibitem[{Scherer and Wallbott(1994)}]{scherer1994evidence}
Klaus~R Scherer and Harald~G Wallbott. 1994.
\newblock {Evidence for Universality and Cultural Variation of Differential
  Emotion Response Patterning}.
\newblock \emph{Journal of personality and social psychology}, 66(2):310.

\bibitem[{Schuff et~al.(2017)Schuff, Barnes, Mohme, Pad{\'o}, and
  Klinger}]{schuff2017annotation}
Hendrik Schuff, Jeremy Barnes, Julian Mohme, Sebastian Pad{\'o}, and Roman
  Klinger. 2017.
\newblock Annotation, modelling and analysis of fine-grained emotions on a
  stance and sentiment detection corpus.
\newblock In \emph{Proceedings of the 8th Workshop on Computational Approaches
  to Subjectivity, Sentiment and Social Media Analysis}, pages 13--23.

\bibitem[{Strapparava and Mihalcea(2007)}]{affective-text-2007-semeval}
Carlo Strapparava and Rada Mihalcea. 2007.
\newblock \href {https://www.aclweb.org/anthology/S07-1013} {{S}em{E}val-2007
  task 14: Affective text}.
\newblock In \emph{Proceedings of the Fourth International Workshop on Semantic
  Evaluations ({S}em{E}val-2007)}, pages 70--74, Prague, Czech Republic.
  Association for Computational Linguistics.

\bibitem[{Tenney et~al.(2019)Tenney, Das, and Pavlick}]{tenney2019bert}
Ian Tenney, Dipanjan Das, and Ellie Pavlick. 2019.
\newblock \href {https://arxiv.org/abs/1905.05950} {{BERT Rediscovers the
  Classical NLP Pipeline}}.
\newblock In \emph{Association for Computational Linguistics}.

\bibitem[{Tsai et~al.(2019)Tsai, Riesa, Johnson, Arivazhagan, Li, and
  Archer}]{tsai2019small}
Henry Tsai, Jason Riesa, Melvin Johnson, Naveen Arivazhagan, Xin Li, and Amelia
  Archer. 2019.
\newblock {Small and Practical BERT Models for Sequence Labeling}.
\newblock In \emph{EMNLP 2019}.

\bibitem[{Wang et~al.(2012)Wang, Chen, Thirunarayan, and
  Sheth}]{wang2012harnessing}
Wenbo Wang, Lu~Chen, Krishnaprasad Thirunarayan, and Amit~P Sheth. 2012.
\newblock Harnessing twitter `big data' for automatic emotion identification.
\newblock In \emph{2012 International Conference on Privacy, Security, Risk and
  Trust and 2012 International Confernece on Social Computing}, pages 587--592.
  IEEE.

\bibitem[{Yanardag and Rahwan(2018)}]{Yanardag2019norman}
Cebrian-M. Yanardag, P. and I.~Rahwan. 2018.
\newblock \href {"http://norman-ai.mit.edu/"} {Norman: World’s first
  psychopath ai}.

\end{thebibliography}
\bibliographystyle{acl_natbib}

\clearpage
\appendix

\section{Emotion Definitions}
\label{sec:emotion_definitions}
\noindent{\bf admiration \includegraphics[height=1em]{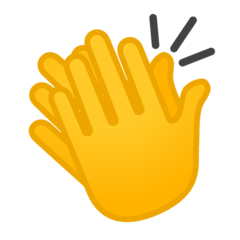}} Finding something impressive or worthy of respect.\\
\noindent{\bf amusement \includegraphics[height=1em]{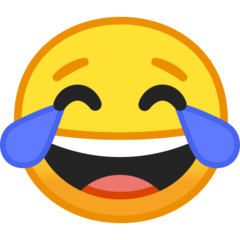}}
Finding something funny or being entertained.\\
\noindent{\bf anger \includegraphics[height=1em]{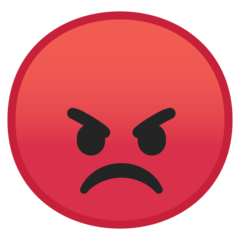}}
A strong feeling of displeasure or antagonism.\\
\noindent{\bf annoyance \includegraphics[height=1em]{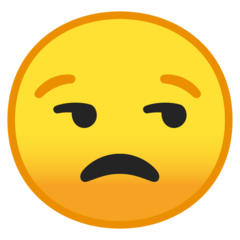}}
Mild anger, irritation.\\
\noindent{\bf approval \includegraphics[height=1em]{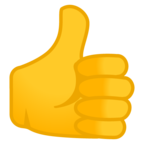}}
Having or expressing a favorable opinion.\\
\noindent{\bf caring}
Displaying kindness and concern for others.\\
\noindent{\bf confusion \includegraphics[height=1em]{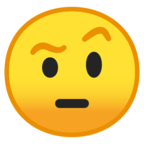}}
Lack of understanding, uncertainty.\\
\noindent{\bf curiosity}
A strong desire to know or learn something.\\
\noindent{\bf desire}
A strong feeling of wanting something or wishing for something to happen.\\
\noindent{\bf disappointment}
Sadness or displeasure caused by the nonfulfillment of one's hopes or expectations.
\noindent{\bf disapproval \includegraphics[height=1em]{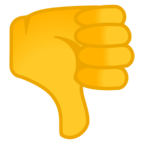}}
Having or expressing an unfavorable opinion.\\
\noindent{\bf disgust \includegraphics[height=1em]{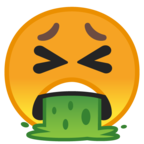}}
Revulsion or strong disapproval aroused by something unpleasant or offensive.\\
\noindent{\bf embarrassment \includegraphics[height=1em]{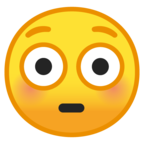}}
Self-consciousness, shame, or awkwardness.\\
\noindent{\bf excitement \includegraphics[height=1em]{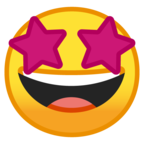}}
Feeling of great enthusiasm and eagerness.\\
\noindent{\bf fear \includegraphics[height=1em]{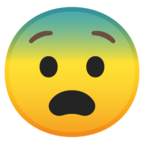}}
Being afraid or worried.\\
\noindent{\bf gratitude \includegraphics[height=1em]{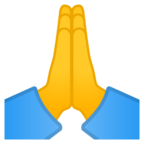}}
A feeling of thankfulness and appreciation.\\
\noindent{\bf grief}
Intense sorrow, especially caused by someone's death.\\
\noindent{\bf joy \includegraphics[height=1em]{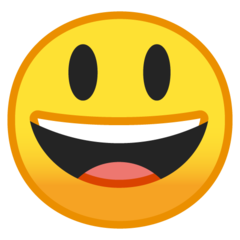}}
A feeling of pleasure and happiness.\\
\noindent{\bf love \includegraphics[height=1em]{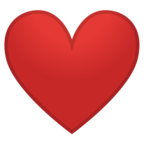}}
A strong positive emotion of regard and affection.\\
\noindent{\bf nervousness}
Apprehension, worry, anxiety.\\
\noindent{\bf optimism \includegraphics[height=1em]{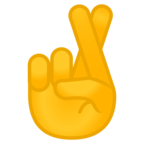}}
Hopefulness and confidence about the future or the success of something.\\
\noindent{\bf pride}
Pleasure or satisfaction due to ones own achievements or the achievements of those with whom one is closely associated.\\
\noindent {\bf realization}
Becoming aware of something.\\
\noindent {\bf relief}
Reassurance and relaxation following release from anxiety or distress.\\
\noindent {\bf remorse}
Regret or guilty feeling.\\
\noindent{\bf sadness \includegraphics[height=1em]{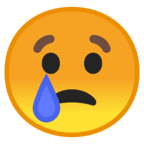}}
Emotional pain, sorrow.\\
\noindent{\bf surprise \includegraphics[height=1em]{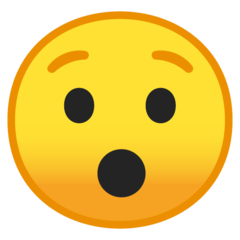}}
Feeling astonished, startled by something unexpected.



\section{Taxonomy Selection \& Data Collection}
\label{sec:appendix_taxonomy_selection}
We selected our taxonomy through a careful multi-round process. In the first pilot round of data collection, we used emotions that were identified to be salient by \citet{cowen2017self}, making sure that our set includes Ekman’s emotion categories, as used in previous NLP work. In this round, we also included an open input box where annotators could suggest emotion(s) that were not among the options. We annotated 3K examples in the first round. We updated the taxonomy based on the results of this round (see details below). In the second pilot round of data collection, we repeated this process with 2k new examples, once again updating the taxonomy.

While reviewing the results from the pilot rounds, we identified and removed emotions that were scarcely selected by annotators and/or had low interrater agreement due to being very similar to other emotions or being too difficult to detect from text. These emotions were \emph{boredom}, \emph{doubt}, \emph{heartbroken}, \emph{indifference} and \emph{calmness}. We also identified and added those emotions to our taxonomy that were frequently suggested by raters and/or seemed to be represented in the data upon manual inspection. These emotions were \emph{desire}, \emph{disappointment}, \emph{pride}, \emph{realization}, \emph{relief} and \emph{remorse}. In this process, we also refined the category names (e.g. replacing \emph{ecstasy} with \emph{excitement}), to ones that seemed interpretable to annotators. This is how we arrived at the final set of 27 emotions + Neutral. Our high interrater agreement in the final data can be partially explained by the fact that we took interpretability into consideration while constructing the taxonomy. The dataset is we are releasing was labeled in the third round over the final taxonomy. 

\section{Cohen's Kappa Values}
\label{sec:appendix_kappa}

In Section~\ref{ssec:interrater_corr}, we measure agreement between raters via Spearman correlation, following considerations by \citet{delgado2019cohen}. In Table~\ref{tab:emo_agreements}, we report the Cohen's kappa values for comparison, which we obtain by randomly sampling two ratings for each example and calculating the Cohen's kappa between these two sets of ratings. We find that all Cohen's kappa values are greater than 0, showing rater agreement. Moreover, the Cohen's kappa values correlate highly with the interrater correlation values (Pearson $r=0.85$, $p < 0.001$), providing corroborative evidence for the significant degree of interrater agreement for each emotion. 

\begin{table}[h!]
    \centering
    \resizebox{.9\linewidth}{!}{
    \begin{centering}
\begin{tabular}{@{}lcc@{}}
\toprule
\textbf{Emotion} & \textbf{\begin{tabular}[c]{@{}c@{}}Interrater\\ Correlation\end{tabular}} & \textbf{\begin{tabular}[c]{@{}c@{}}Cohen's\\ kappa\end{tabular}} \\ \midrule
admiration       & 0.535                                                                     & 0.468                                                            \\
amusement        & 0.482                                                                     & 0.474                                                            \\
anger            & 0.207                                                                     & 0.307                                                            \\
annoyance        & 0.193                                                                     & 0.192                                                            \\
approval         & 0.385                                                                     & 0.187                                                            \\
caring           & 0.237                                                                     & 0.252                                                            \\
confusion        & 0.217                                                                     & 0.270                                                            \\
curiosity        & 0.418                                                                     & 0.366                                                            \\
desire           & 0.177                                                                     & 0.251                                                            \\
disappointment   & 0.186                                                                     & 0.184                                                            \\
disapproval      & 0.274                                                                     & 0.234                                                            \\
disgust          & 0.192                                                                     & 0.241                                                            \\
embarrassment    & 0.177                                                                     & 0.218                                                            \\
excitement       & 0.193                                                                     & 0.222                                                            \\
fear             & 0.266                                                                     & 0.394                                                            \\
gratitude        & 0.645                                                                     & 0.749                                                            \\
grief            & 0.162                                                                     & 0.095                                                            \\
joy              & 0.296                                                                     & 0.301                                                            \\
love             & 0.446                                                                     & 0.555                                                            \\
nervousness      & 0.164                                                                     & 0.144                                                            \\
optimism         & 0.322                                                                     & 0.300                                                            \\
pride            & 0.163                                                                     & 0.148                                                            \\
realization      & 0.194                                                                     & 0.155                                                            \\
relief           & 0.172                                                                     & 0.185                                                            \\
remorse          & 0.178                                                                     & 0.358                                                            \\
sadness          & 0.346                                                                     & 0.336                                                            \\
surprise         & 0.275                                                                     & 0.331                                                            \\ \bottomrule
\end{tabular}
    \end{centering}
    }
    \caption{Interrater agreement, as measured by interrater correlation and Cohen's kappa}
    \label{tab:emo_agreements}
\end{table}

\section{Sentiment of Reddit Subreddits}
In Section~\ref{sec:data_collection}, we describe how we obtain subreddits that are balanced in terms of sentiment. Here, we note the distribution of sentiments across subreddits \emph{before} we apply the filtering: neutral (M=28\%, STD=11\%), positive (M=41\%, STD=11\%), negative (M=19\%, STD=7\%), ambiguous (M=35\%, STD=8\%). After filtering, the distribution of sentiments across our remaining subreddits became: neutral (M=24\%, STD=5\%), positive (M=35\%, STD=6\%), negative (M=27\%, STD=4\%), ambiguous (M=33\%, STD=4\%).

\section{BERT's Most Activated Layers}

\begin{figure}[h]
 \centering
   \centering
   \includegraphics[width=\linewidth]{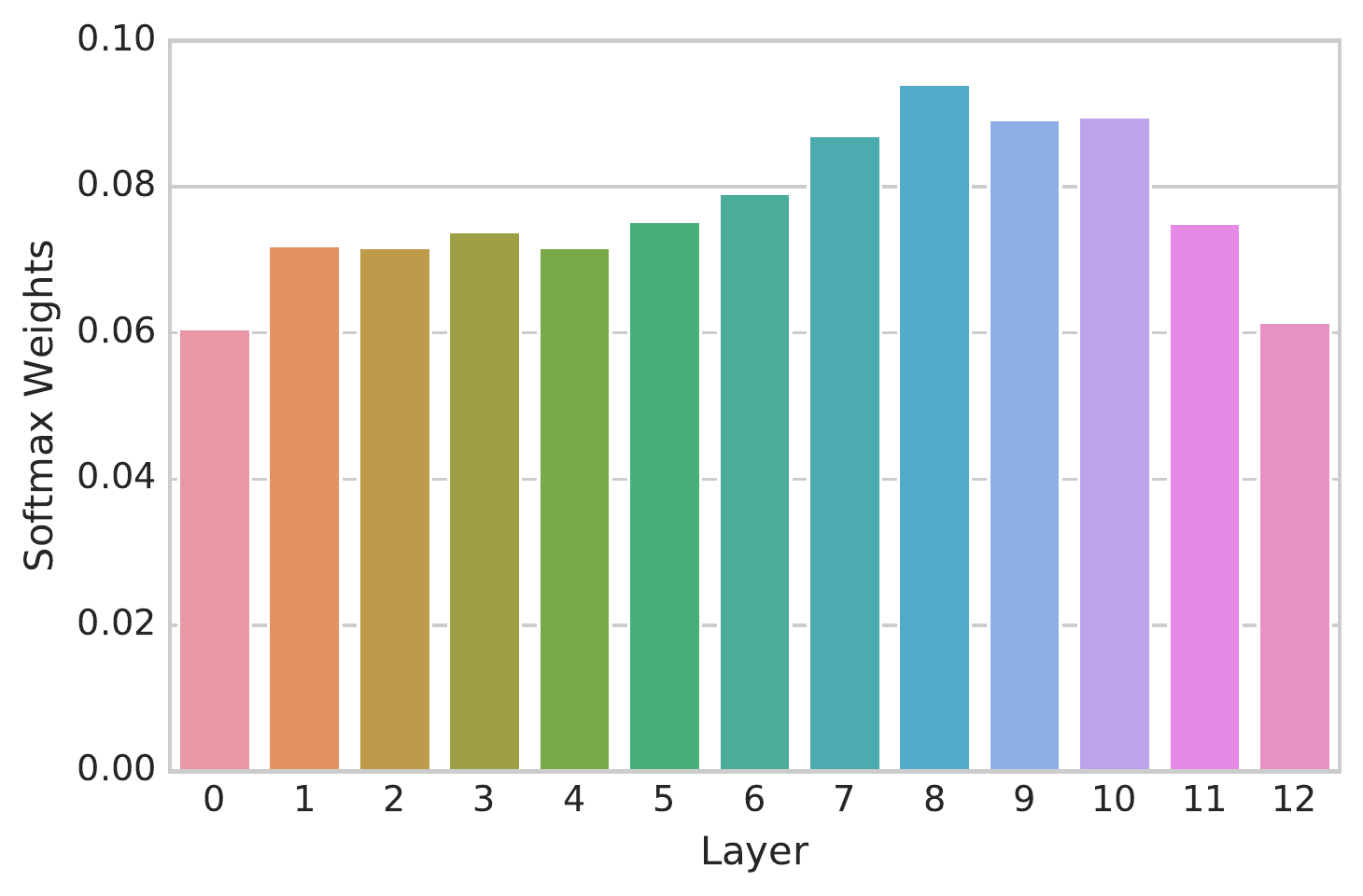}
   \caption{Softmax weights of each BERT layer when trained on our dataset.}
   \label{fig:bert_gravity}
\end{figure}

To better understand whether there are any layers in BERT that are particularly important for our task, we freeze BERT and calculate the center of gravity \citep{tenney2019bert} based on scalar mixing weights \citep{peters2018deep}. We find that all layers are similarly important for our task, with center of gravity = 6.19 (see Figure~\ref{fig:bert_gravity}). This is consistent with \citet{tenney2019bert}, who have also found that tasks involving high-level semantics tend to make use of all BERT layers.

\section{Number of Emotion Labels Per Example}
\label{sec:modeling_plots}

\begin{figure}[h]
 \centering
   \centering
   \includegraphics[width=\linewidth]{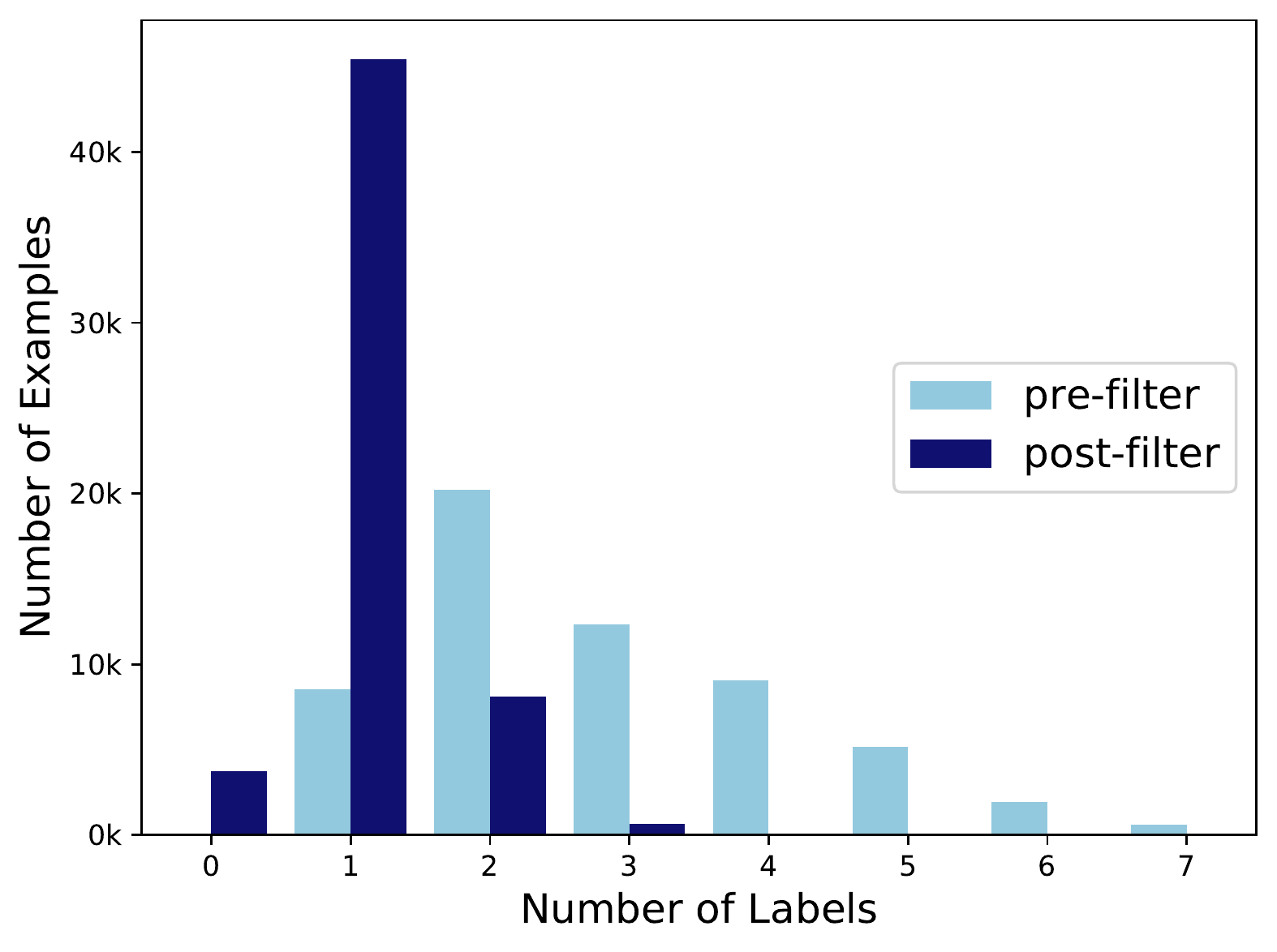}
   \caption{Number of emotion labels per example before and after filtering the labels chosen by only a single annotator.}
   \label{fig:num_labels}
\end{figure}

Figure~\ref{fig:num_labels} shows the number of emotion labels per example before and after we filter for those labels that have agreement. We use the filtered set of labels for training and testing our models.

\section{Confusion Matrix}
\label{sec:appendix_confusion_matrix}

\begin{figure*}[t!]
 \centering
   \centering
   \includegraphics[width=.7\linewidth]{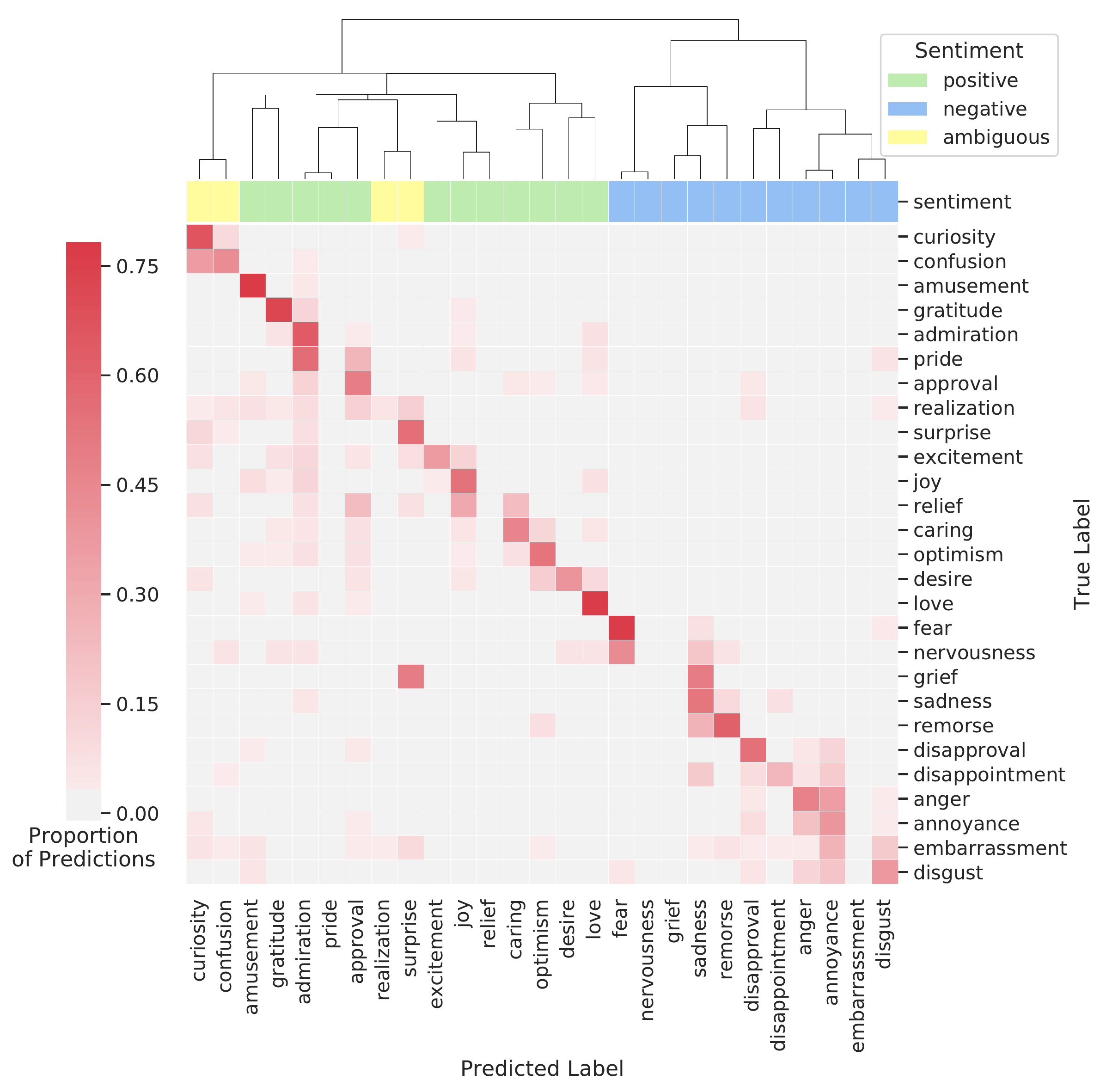}
   \caption{A normalized confusion matrix for our model predictions. The plot shows that the model confuses emotions with other emotions that are related in intensity and sentiment.}
   \label{fig:confusion_matrix}
\end{figure*}

Figure~\ref{fig:confusion_matrix} shows the normalized confusion matrix for our model predictions. Since GoEmotions is a multilabel dataset, we calculate the confusion matrix similarly as we would calculate a co-occurrence matrix: for each true label, we increase the count for each predicted label. Specifically, we define a matrix $M$ where $M_{i, j}$ denotes the raw confusion count between the true label $i$ and the predicted label $j$. For example, if the true labels are \emph{joy} and \emph{admiration}, and the predicted labels are \emph{joy} and \emph{pride}, then we increase the count for $M_{joy, joy}$, $M_{joy, pride}$, $M_{admiration, joy}$ and $M_{admiration, pride}$. In practice, since most of our examples only has a single label (see Figure~\ref{fig:num_labels}), our confusion matrix is very similar to one calculated for a single-label classification task.

Given the disparate frequencies among the labels, we normalize $M$ by dividing the counts in each row (representing counts for each true emotion label) by the sum of that row. The heatmap in Figure~\ref{fig:confusion_matrix} shows these normalized counts. We find that the model tends to confuse emotions that are related in sentiment and intensity (e.g., \emph{grief} and \emph{sadness}, \emph{pride} and \emph{admiration}, \emph{nervousness} and \emph{fear}).

We also perform hierarchical clustering over the normalized confusion matrix using correlation as a distance metric and ward as a linkage method. We find that the model learns relatively similar clusters as the ones in Figure~\ref{fig:hierarchical_corr}, even though the training data only includes a subset of the labels that have agreement (see Figure~\ref{fig:num_labels}).

\section{Transfer Learning Results}
\label{sec:appendix_transfer_learning}

\begin{figure*}[t!]
 \centering
   \centering
   \includegraphics[width=\linewidth]{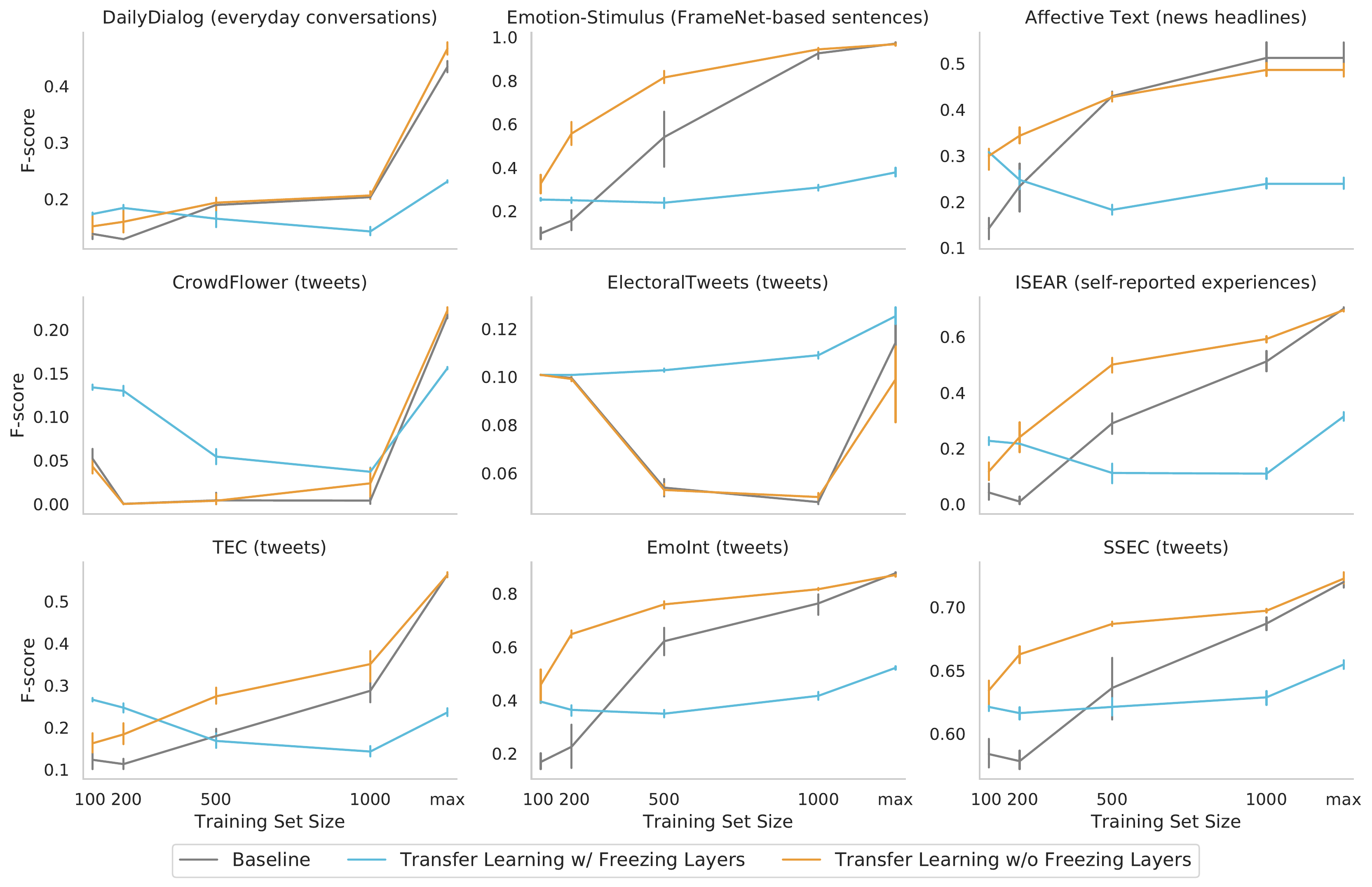}
   \caption{Transfer learning results on 9 emotion benchmarks from the Unified Dataset \citep{unified-klinger2018analysis}.}
   \label{fig:transfer_learning_all}
\end{figure*}

Figure~\ref{fig:transfer_learning_all} shows the results for all 9 datasets that are downloadable and have categorical emotions in the Unified Dataset \citep{unified-klinger2018analysis}. These datasets are DailyDialog \citep{li2017dailydialog}, Emotion-Stimulus \citep{ghazi2015detecting}, Affective Text \citep{affective-text-2007-semeval}, CrowdFlower \citep{crowdflower-2016}, Electoral Tweets \citep{electoraltweets-mohammad2015sentiment}, ISEAR \citep{scherer1994evidence}, the Twitter Emotion Corpus (TEC) \citep{tec-mohammad2012emotional}, EmoInt \citep{mohammad-etal-2018-semeval} and the Stance Sentiment Emotion Corpus (SSEC) \citep{schuff2017annotation}. 

We describe the experimental setup in Section~\ref{ssec:transfer_learning_setup}, which we use across all datasets. We find that transfer learning helps in the case of all datasets, especially when there is limited training data. Interestingly, in the case of CrowdFlower, which is known to be noisy \citep{unified-klinger2018analysis} and Electoral Tweets, which is a small dataset of $\sim$4k labeled examples and a large taxonomy of 36 emotions, \textsc{freeze} gives a significant boost of performance over the \textsc{baseline} and \textsc{nofreeze} for all training set sizes besides ``max''. 

For the other datasets, we find that \textsc{freeze} tends to give a performance boost compared to the other setups only up to a couple of hundred training examples. For 500-1000 training examples, \textsc{nofreeze} tends to outperform the \textsc{baseline}, but we can see that these two setups come closer when there is more training data available. These results suggests that our dataset helps if there is limited data from the target domain.

\end{document}